\pdfoutput=1
\documentclass{article}
\usepackage[nusbwlab]{iclr2025_conference}
\usepackage{iftex}
\ifPDFTeX
  \usepackage[T1]{fontenc}
  \usepackage{times}
\else
  \usepackage{fontspec}
\fi
\bwlheadplain
\usepackage{amsmath,amssymb,booktabs,graphicx,microtype,hyperref}
\usepackage{adjustbox,multirow,flafter,needspace,placeins}
\usepackage{colortbl,siunitx}
\definecolor{maintableblue}{HTML}{E4ECF8}


\renewenvironment{abstract}{%
  \par\smallskip
  \centerline{\sffamily\bfseries\large\color{bwlblue}Abstract}%
  \begin{list}{}{\setlength{\leftmargin}{1.5em}%
    \setlength{\rightmargin}{1.5em}\setlength{\topsep}{5pt}}%
  \item\relax
}{\end{list}}

\graphicspath{{./}}
\title{MolReJev: Decision Models Are \mbox{In-Context Molecular Predictors}}
\author{%
Kaiyuan Lu\textsuperscript{*}\qquad
Jiatong Li\textsuperscript{*}\qquad
Yatao Bian\textsuperscript{\(\dagger\)}\\
National University of Singapore\\[2pt]
{\normalfont\footnotesize \textsuperscript{*}Equal contribution.\quad
\textsuperscript{\(\dagger\)}Corresponding author: \texttt{ybian@nus.edu.sg}.}%
}
\begin{document}
\maketitle
\suppressfloats[t]
\begin{abstract}
Molecular prediction helps prioritize compounds for screening. Specialized predictors are trained for a fixed set of tasks, limiting cross-task generalization. Language models can condition on task descriptions and have recently shown promise in molecular prediction. However, generating answers token by token adds inference cost, and general-purpose models often lack task-specific molecular knowledge. We thus propose MolReJev, which combines Jev-style candidate scoring with in-context learning (ICL) on a language model. MolReJev retrieves similar labeled examples from existing task data and scores candidate labels jointly in one forward pass. ICL improves predictions from the released general-purpose Kev checkpoint and the separately trained MolReJev. For the trained model, gains span all evaluation groups, including held-out tasks, and extend to compounds absent from training. MolReJev generally outperforms MolecularGPT on matched evaluations and nearest-neighbor prediction with the same examples. With four examples, MolReJev and a trained text baseline achieve similar ROC-AUC, with higher measured throughput for MolReJev. These results show that decision models can serve as in-context molecular predictors for compounds, drug pairs, and reactions.
\end{abstract}

\section{Introduction}
A compound that inhibits one protein can be inactive against another. Activity predictions help prioritize compounds for a particular screen, and properties such as blood--brain barrier permeability inform which candidates to advance. Changes in the assay or target can reduce the accuracy of predictors trained on existing data \citep{ji2023drugood}. For a new target with only a handful of labeled compounds, the challenge is to use those labels to rank the rest of the library \citep{stanley2021fs}.

Language models can condition predictions on task descriptions. Controlled perturbation studies show that even small structural edits can affect their performance on molecular tasks \citep{li2026llms}. In-context learning (ICL) provides labeled examples of how related molecules behave on the task, with the model weights held fixed. MolReGPT retrieves similar molecule--caption pairs for translation \citep{li2024empowering}, and MolecularGPT uses structurally related labeled compounds for property prediction \citep{liu2024moleculargpt}.

Text-based prediction requires sequential decoding. Scoring a candidate label that spans multiple tokens also requires evaluating its sequence likelihood. Jev-style decision models score supplied labels jointly and return a probability distribution in one forward pass \citep{almeida2026jev}. We therefore ask whether decision models can learn from labeled molecular examples in context and use them to make predictions on new tasks.

We propose MolReJev, which combines Jev-style candidate scoring with ICL on a language model. For each query, MolReJev retrieves similar labeled examples from existing data for the task and scores the candidate labels in light of those examples. MolReJev (Untrained) uses the released general-purpose Kev checkpoint\footnote{\url{https://github.com/jaredpalmer/kev}} with its weights fixed. For MolReJev (Trained), we train a separate decision model on Qwen3-4B across molecular tasks using the same prompt format.

\begin{figure}[!htbp]
\centering
\includegraphics[width=\linewidth]{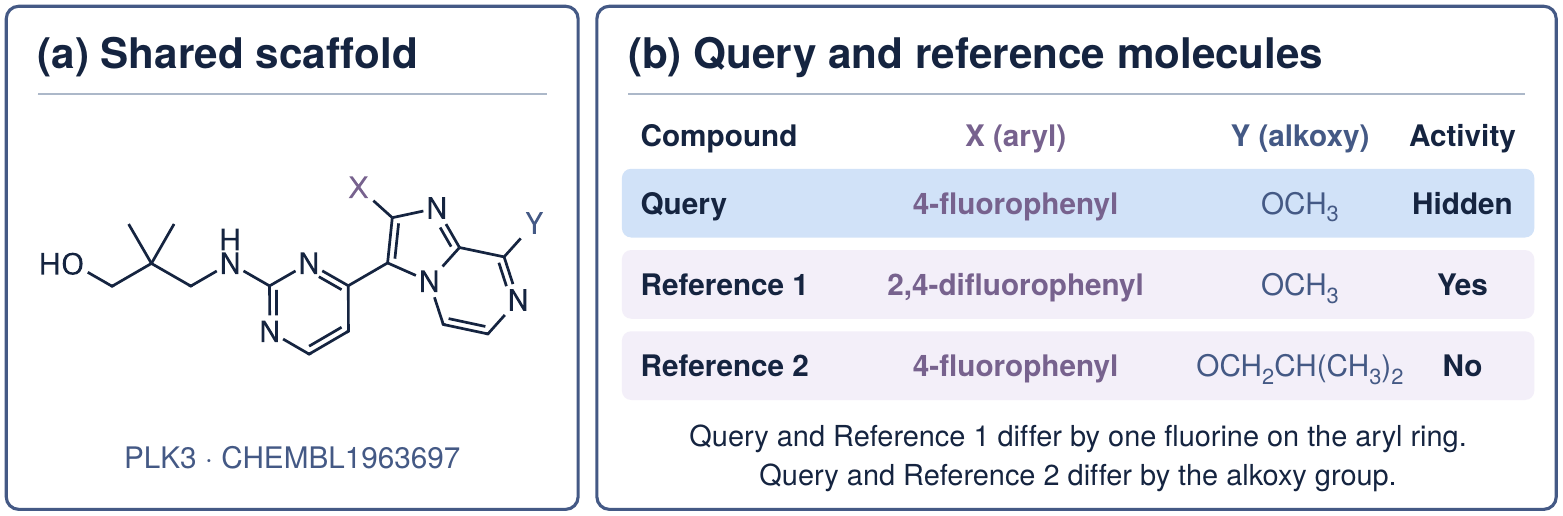}
\caption{PLK3 evaluation query and two of its four retrieved references from FS-Mol assay CHEMBL1963697 \citep{stanley2021fs,mendez2019chembl}. The X and Y substituents specify the compounds on the shared scaffold. The query label is hidden.}
\label{fig:intro}
\end{figure}

Figure~\ref{fig:intro} shows an evaluation query from an FS-Mol assay for polo-like kinase 3 (PLK3). Two of the four retrieved references are shown. They share the query's scaffold and carry different labels. The active reference has one additional fluorine on the aryl ring, and the inactive reference differs in its alkoxy group. MolReJev uses the retrieved structures and their labels to score the query.
\FloatBarrier

Retrieved examples improve predictions from the released Kev checkpoint and from the separately trained model. The trained model improves in all 18 evaluation groups, with gains on held-out tasks and compounds absent from molecular training. MolReJev outperforms MolecularGPT on most matched evaluations and usually exceeds nearest-neighbor prediction given the same examples. The trained text baseline also benefits from ICL. With four examples, MolReJev achieves comparable predictive performance and higher measured throughput. These results support decision models as in-context predictors for compounds, drug pairs, and reactions.

\section{Related Work}
\paragraph{Decision models.} Jev assigns probabilities to options supplied in the input \citep{almeida2026jev}. Kev implements this interface on open-weight backbones, using a pointer head to score option positions \citep{vinyals2015pointer}. Jev has been evaluated on text classification, routing, and content moderation \citep{deusser2026evaluating}. \citet{lyu2026decision} use it to rerank candidate items in personalized recommendation. In medicine, \citet{du2026reliable} evaluate decision accuracy, consistency under changes in question format, and the ability to abstain.

\paragraph{In-context learning across molecular tasks.} Language models can infer a task from examples in the prompt \citep{brown2020language}, and the choice of examples affects prediction accuracy \citep{rubin2022learning,liu2022makes}. Training on demonstrations across tasks can improve transfer \citep{min2022metaicl,chen2022meta}, and instruction tuning specifies tasks through natural-language descriptions \citep{wei2021finetuned,sanh2021multitask}. In molecular prediction, FS-Mol benchmarks transfer to new assays using labeled support molecules \citep{stanley2021fs}. Nearest-neighbor methods use labels of similar compounds, prototypical networks compare queries with class representations \citep{snell2017prototypical}, and property-aware relation networks learn how compounds relate under a given property \citep{wang2021property}. MolReGPT and ICMA retrieve examples for molecule--text translation \citep{li2024empowering,li2025large}, and MolReFlect retrieves prior alignments to refine links between substructures and caption phrases \citep{li2026molreflect}. MolecularGPT combines instruction tuning with retrieved examples for property prediction \citep{liu2024moleculargpt}. Analyses of chemical language models probe the faithfulness of generated reasoning and whether latent states act as soft prompts \citep{li2026chemical,lu2026reasoning}.

\section{MolReJev}
\subsection{Problem formulation}
A task $t$ supplies a description $a_t$, candidate labels $Y_t$, and a labeled reference pool $P_t=\{(x_i,y_i)\}$ with $y_i\in Y_t$. Given a query $x$, we score each candidate label. The PLK3 assay in Figure~\ref{fig:intro} defines two labels, DrugBank defines 86 interaction types, and USPTO defines 50 reaction classes. Queries are single molecules, drug pairs, or reactions. Small molecules and reactions use SMILES \citep{weininger1988smiles}, and antibodies use amino acid sequences. Binding tasks also include the target sequence.

\subsection{Context retrieval and prompt construction}
For each query, we rank references in $P_t$ by their similarity scores and take the top $k$. For small molecules, we use Tanimoto similarity on MACCS fingerprints \citep{durant2002reoptimization,bajusz2015tanimoto}. For drug pairs, we average the Tanimoto scores of the two drugs matched by position. We use cosine similarity on TF--IDF vectors of amino acid trigrams for antibodies and character trigrams of reaction SMILES for reactions. Ties follow reference-pool order (Table~\ref{tab:retrieval-protocol}). Training queries exclude themselves, and evaluation queries are disjoint from the reference pool for their task.

\begin{figure}[!htbp]
\centering
\includegraphics[width=\linewidth]{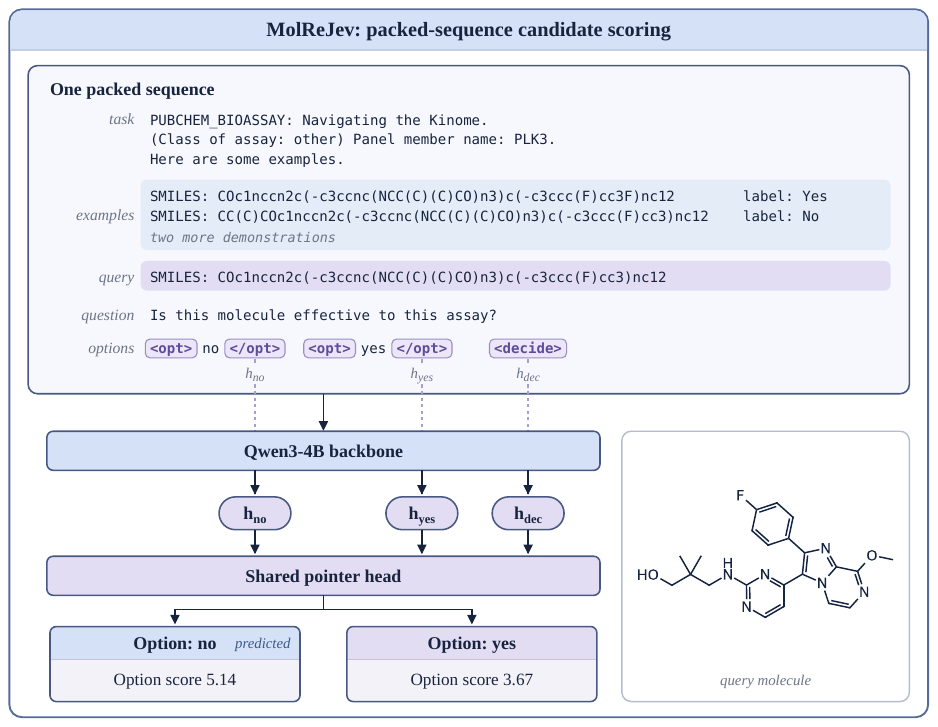}
\caption{Candidate scoring for the PLK3 query in Figure~\ref{fig:intro}. The diagram shows two of the four examples in the packed input. Trained MolReJev assigns scores of 5.14 to \texttt{no} and 3.67 to \texttt{yes}, predicting \texttt{no}, the recorded label.}
\label{fig:method}
\end{figure}

The prompt starts with the task description and the target sequence when available. We then list the retrieved examples and their labels in decreasing order of similarity and append the unlabeled query. We use the final sentence of $a_t$ as the question. The candidate labels follow as separate delimited options, and the prompt ends with a \texttt{<decide>} token. Figure~\ref{fig:method} shows the prompt and scores for the PLK3 query. Appendix~\ref{app:details} gives the template.

\subsection{Candidate scoring}\label{sec:scoring}
MolReJev uses a pointer head based on Kev. MolReJev (Untrained) runs the released Kev checkpoint with the retrieved examples in context, keeping its weights fixed. For the trained setting, we use the Qwen3-4B backbone \citep{yang2025qwen3} and train a randomly initialized pointer head alongside the backbone adapters on molecular records. For each label $y \in Y_t$, the model reads the final-layer hidden state $h_y \in \mathbb{R}^d$ at the label's closing delimiter and compares it with $h_{\mathrm{dec}} \in \mathbb{R}^d$, the hidden state at \texttt{<decide>}. The projections $W_q, W_k \in \mathbb{R}^{d_p \times d}$ are shared across labels, with $d_p = 256$. Given the retrieved examples $D_k(x)$, the score and probability of label $y$ are

\begin{equation}z_y=\frac{(W_q h_{\mathrm{dec}})^\top (W_k h_y)}{\sqrt{d_p}}, \qquad p(y \mid x, a_t, D_k(x), Y_t)=\frac{\exp z_y}{\sum_{y' \in Y_t} \exp z_{y'}}. \end{equation}

Training minimizes the negative log probability of the observed label.

One pass scores every member of $Y_t$. Within a binary task we rank queries by $z_{\mathrm{yes}} - z_{\mathrm{no}}$, which orders them by the probability of yes; for the multi-class tasks we take the label of highest score as the prediction. Larger candidate sets lengthen the input, but all labels are scored in a single forward pass.

\subsection{Cross-task training}\label{sec:training}
Each training record uses one labeled pool member as the query and its four nearest remaining neighbours from the same task as demonstrations. We construct 489,116 such records from the molecular tasks (440,116), DrugBank DDI (40,000), and USPTO (9,000) (Table~\ref{tab:data}). We train for one epoch and report one run per trained system in the main results.

MolReJev and the text baseline of Appendix~\ref{app:generative} see the same records and share their optimization settings: rank-16 LoRA adapters \citep{hu2021lora} ($\alpha=32$, dropout 0.05) on all linear projections, AdamW \citep{loshchilov2017decoupled} with weight decay 0.01, a one-cycle schedule with a peak learning rate of $5\times10^{-5}$ chosen on the molecular records for both heads, an effective batch of eight records, and bfloat16 precision. Two details follow the format rather than the optimizer. MolReJev caps a record at 3,072 tokens and the text baseline at 4,096, which leaves out 120 and 12 records respectively. MolReJev trains a pointer head of width 256 alongside its adapters, two $2560\times256$ projections with biases and 1.31M parameters in all; the text baseline updates only its adapters, as its language-model head shares weights with the input embeddings and stays frozen.

We call the language-model backbone before training Qwen3-4B and the two trained systems MolReJev (Trained) and Qwen3-4B-Text. We use MolReJev for the trained setting unless specified. By default, we supply four retrieved examples ($k=4$). Kev denotes the released checkpoint at $k=0$, and MolReJev (w/o ICL) denotes the trained weights at $k=0$. For held-out tasks, we retrieve examples from each task's own pool with the weights held fixed.

\section{Experimental Settings}
We evaluate two settings of the MolReJev framework. MolReJev (Untrained) uses the released Kev checkpoint, and MolReJev uses the model trained across molecular tasks. Section~\ref{sec:training} gives the training details and Appendix~\ref{app:details} lists the configurations.

\paragraph{Datasets and splits.} We evaluate small-molecule properties, compound--protein interactions, antibody binding, drug--drug interaction and reaction classification in 18 groups (Table~\ref{tab:data}). Ten groups test new queries from tasks seen during training. The other eight test tasks excluded from that training. FS-Mol contributes both types: its 4,507 training assays and its 157 held-out assays form separate groups. AVIDa-hIL6 likewise contributes two groups split by antigen variant. A single dataset may contain several tasks; for example, Tox21 has 12 toxicity endpoints. Appendix~\ref{app:details} gives the split rules and query counts.

\paragraph{Baselines.} Qwen3-4B uses its original language-model head, scoring binary queries by the Yes--No logit difference and multiclass queries by candidate-string log-likelihood. Qwen3-4B-Text receives the same prompt content as MolReJev, lists the candidate labels one per line, and generates a JSON answer. We score binary queries with the Yes--No logit difference (Appendix~\ref{app:generative}). We also evaluate the released MolecularGPT adapter \citep{liu2024moleculargpt} in its original prompt format on ten binary and two multiclass groups that require neither protein nor antibody sequences.

The $k$-nearest-neighbor predictor uses the same four references, weighting their labels by similarity for binary tasks and using majority voting for multiclass tasks. We fit a random forest per task on 16, 64, or all labeled reference records, capped by the pool size (Table~\ref{tab:rf-full}). On held-out FS-Mol assays, we also evaluate the released ProtoNet checkpoint with randomly sampled support molecules (Appendix~\ref{app:details}).

\Needspace{6\baselineskip}
\paragraph{Evaluation metrics.} For a task $t$ with positive and negative query sets $Q_t^+$ and $Q_t^-$, ROC-AUC measures how often a positive query scores above a negative query \citep{hanley1982meaning}. For query scores $s_i$, it gives half credit to ties:
\[
\mathrm{ROC\text{-}AUC}_t =
\frac{\sum_{i\in Q_t^+}\sum_{j\in Q_t^-}
\bigl(\mathbf{1}[s_i>s_j]+\tfrac12\mathbf{1}[s_i=s_j]\bigr)}
{|Q_t^+|\,|Q_t^-|}.
\]
We average over tasks with both labels within each group, then weight all 16 ROC-AUC groups equally. For a multiclass task with $n$ queries and classes $\mathcal{C}$ present in the evaluation labels, we report
\[
\mathrm{Accuracy}=\frac{1}{n}\sum_{i=1}^{n}\mathbf{1}[\hat y_i=y_i],
\qquad
\mathrm{Macro\text{-}F1}=\frac{1}{|\mathcal{C}|}\sum_{c\in\mathcal{C}}
\frac{2\mathrm{TP}_c}{2\mathrm{TP}_c+\mathrm{FP}_c+\mathrm{FN}_c}.
\]
Here $y_i$ and $\hat y_i$ are the true and predicted labels, and $\mathrm{TP}_c$, $\mathrm{FP}_c$, and $\mathrm{FN}_c$ count true positives, false positives, and false negatives for class $c$.

\Needspace{5\baselineskip}
\section{Results}\label{sec:results}

\subsection{In-context learning in a decision model}\label{sec:icl}
We test how much four retrieved examples improve decision models at fixed weights and compare trained MolReJev with Qwen3-4B given four examples each. Table~\ref{tab:main} shows the group-level results for Qwen3-4B and trained MolReJev at both example counts. Table~\ref{tab:components} summarizes the fixed-weight Kev comparison and the four-example comparison using mean ROC-AUC.
\begin{table}[!htbp]
\centering\small
\caption{Molecular prediction with $k$ retrieved examples. ROC-AUC is averaged over tasks within each group. Mean rows weight the indicated groups equally. Bold marks the best score in each row.}
\label{tab:main}
\sisetup{table-format=1.3, mode=text, reset-text-series=false,
  table-fixed-width=true, table-column-width=.14\linewidth}
\setlength{\tabcolsep}{5pt}
\begin{tabular*}{\linewidth}{@{\extracolsep{\fill}}l S S S S@{}}
\toprule
 & \multicolumn{2}{c}{\textbf{Qwen3-4B}} & \multicolumn{2}{c}{\textbf{MolReJev}} \\
\cmidrule(lr){2-3} \cmidrule(lr){4-5}
\textbf{Dataset group} & \multicolumn{1}{c}{$k=0$} & \multicolumn{1}{c}{$k=4$}
 & \multicolumn{1}{c}{$k=0$} & \multicolumn{1}{c}{\cellcolor{maintableblue}$k=4$} \\
\midrule
\rowcolor{maintableblue!65}[0pt][\tabcolsep]
\multicolumn{5}{@{}l}{\textit{ROC-AUC (seen in molecular training)}} \\
FS-Mol (4507)              & 0.481 & 0.577 & 0.644 & \bfseries 0.782 \\
BACE (1)                   & 0.421 & 0.598 & 0.600 & \bfseries 0.861 \\
BBBP (1)                   & 0.542 & 0.654 & 0.514 & \bfseries 0.694 \\
HIV (1)                    & 0.477 & 0.776 & 0.820 & \bfseries 0.821 \\
SIDER (27)                 & 0.513 & 0.581 & 0.562 & \bfseries 0.640 \\
Tox21 (12)                 & 0.466 & 0.651 & 0.652 & \bfseries 0.751 \\
BioSNAP (35)               & 0.498 & 0.618 & 0.495 & \bfseries 0.797 \\
AVIDa-hIL6 (16)            & 0.540 & 0.909 & 0.558 & \bfseries 0.926 \\
\addlinespace[4pt]
\rowcolor{maintableblue!65}[0pt][\tabcolsep]
\multicolumn{5}{@{}l}{\textit{ROC-AUC (held out from molecular training)}} \\
FS-Mol held-out assays (157)   & 0.491 & 0.566 & 0.635 & \bfseries 0.760 \\
CYP450 (5)                     & 0.523 & 0.804 & 0.593 & \bfseries 0.834 \\
ToxCast (530)                  & 0.487 & 0.556 & 0.561 & \bfseries 0.628 \\
ClinTox (2)                    & 0.461 & 0.673 & 0.536 & \bfseries 0.810 \\
BindingDB (152)                & 0.522 & 0.808 & 0.528 & \bfseries 0.916 \\
AVIDa-hIL6 held-out (15)       & 0.495 & 0.906 & 0.588 & \bfseries 0.944 \\
AVIDa-SARS-CoV-2 (12)          & 0.459 & 0.654 & 0.452 & \bfseries 0.726 \\
AVIDa-hTNF$\alpha$ (1)         & 0.469 & 0.852 & 0.631 & \bfseries 0.893 \\
\midrule
Mean ROC-AUC (8 held-out groups) & 0.488 & 0.727 & 0.565 & \bfseries 0.814 \\
Mean ROC-AUC (16 groups)        & 0.490 & 0.699 & 0.586 & \bfseries 0.799 \\
\addlinespace[4pt]
\rowcolor{maintableblue!65}[0pt][\tabcolsep]
\multicolumn{5}{@{}l}{\textit{Multiclass accuracy}} \\
DrugBank DDI (86 classes)      & 0.005 & 0.778 & 0.361 & \bfseries 0.843 \\
USPTO reactions (50 classes)  & 0.020 & 0.633 & 0.684 & \bfseries 0.912 \\
\addlinespace[4pt]
\rowcolor{maintableblue!65}[0pt][\tabcolsep]
\multicolumn{5}{@{}l}{\textit{Multiclass macro-F1}} \\
DrugBank DDI (86 classes)      & 0.000 & 0.680 & 0.015 & \bfseries 0.779 \\
USPTO reactions (50 classes)  & 0.001 & 0.630 & 0.667 & \bfseries 0.911 \\
\bottomrule
\end{tabular*}
\end{table}

\begin{table}[!htbp]
\centering\small
\caption{Mean ROC-AUC across 16 groups. The first pair tests the effect of retrieved examples at fixed Kev weights. The second compares systems given four examples each.}
\label{tab:components}
\begin{tabular}{@{}llcr@{}}
\toprule
Comparison & System & Examples ($k$) & Mean ROC-AUC \\
\midrule
\multirow{2}{*}{Fixed Kev weights} & Kev & 0 & 0.491 \\
 & MolReJev (Untrained) & 4 & \textbf{0.711} \\
\midrule
\multirow{2}{*}{Four examples each} & Qwen3-4B & 4 & 0.699 \\
 & MolReJev (Trained) & 4 & \textbf{0.799} \\
\bottomrule
\end{tabular}
\end{table}

\Needspace{4\baselineskip}
Four examples raise mean ROC-AUC by 0.220 for Kev and by 0.213 for trained MolReJev. With four examples each, MolReJev reaches 0.799 versus 0.699 for Qwen3-4B. The ICL gains extend to DrugBank DDI and USPTO, where MolReJev chooses among dozens of labels and leads Qwen3-4B in accuracy and macro-F1 at $k=4$.

\subsection{Ablation on in-context examples}\label{sec:demos}
Both ablations use up to the first 32 queries per binary task and 1,000 queries per multiclass task.

\begin{figure}[!htbp]
\centering
\includegraphics[width=\linewidth]{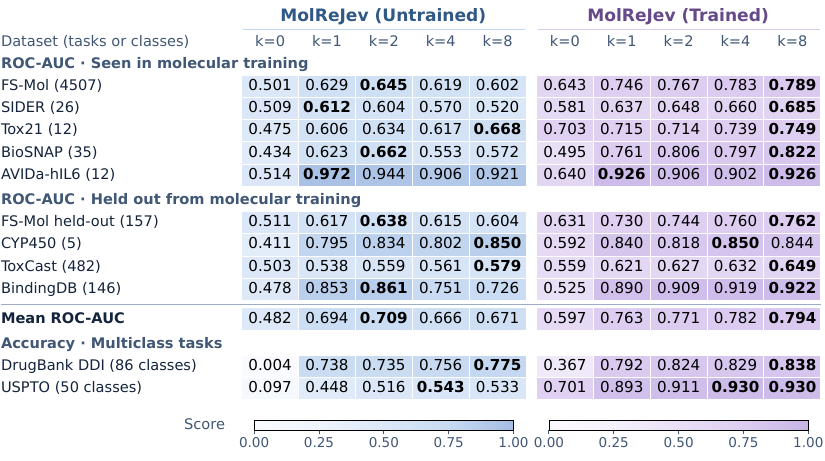}
\caption{Effect of the demonstration count $k$. Binary groups report ROC-AUC on up to 32 queries per task, with evaluable task counts in parentheses. The mean weights the nine groups equally. DrugBank DDI and USPTO report accuracy on 1,000 queries per dataset. Bold marks the best displayed score within each model and row.}
\label{fig:demo-count}
\end{figure}

\paragraph{Number of examples.} We vary the number of examples from zero to eight for both decision model settings on the same queries. Figure~\ref{fig:demo-count} summarizes the sweep over nine ROC-AUC groups and two multiclass datasets.

The first example accounts for most of the gain in mean ROC-AUC for MolReJev (Untrained), raising it from 0.482 to 0.694. It also yields the largest accuracy gain on both multiclass subsamples. The untrained setting's mean peaks at two examples, and eight of nine ROC-AUC groups score lower at four. The trained setting's mean increases at each tested count, reaching 0.794 at eight examples. Individual groups can still decline as more examples are added.

\paragraph{Retrieval strategy.} The retrieval seeds vary the examples while keeping the queries and weights fixed. The four most similar references give the best scores across the groups tested (Table~\ref{tab:retrieval}).

On DrugBank DDI, accuracy falls from 0.829 with the nearest examples to 0.312 with random examples. Class-balanced selection recovers part of the loss from random sampling on AVIDa-hIL6, where positive examples are scarce. A USPTO case illustrates that MolReJev can select a label absent from its retrieved examples (Appendix~\ref{app:cases}).
\begin{table}[!htbp]
\centering\small
\caption{Effect of example selection in MolReJev with four demonstrations. ROC-AUC uses up to 32 queries per task, with evaluable task counts in parentheses. Multiclass metrics use 1,000 queries per dataset. Random and class-balanced sampling report the mean $\pm$ sample standard deviation over four draws. Class-balanced sampling uses two examples per class for binary tasks. Bold marks the best score in each row.}
\label{tab:retrieval}
\sisetup{table-format=1.3, mode=text, reset-text-series=false}
\setlength{\tabcolsep}{5pt}
\begin{tabular*}{\linewidth}{@{\extracolsep{\fill}}l S S[table-format=1.3(3),uncertainty-mode=separate] S[table-format=1.3(3),uncertainty-mode=separate] S@{}}
\toprule
\textbf{Dataset group} & \multicolumn{1}{c}{\cellcolor{maintableblue}\textbf{Most similar}} & \multicolumn{1}{c}{\textbf{Random}} & \multicolumn{1}{c}{\textbf{Balanced}} & \multicolumn{1}{c}{\textbf{Least similar}} \\
\midrule
\rowcolor{maintableblue!65}[0pt][\tabcolsep]
\multicolumn{5}{@{}l}{\textit{ROC-AUC (seen in molecular training)}} \\
FS-Mol (4507)                 & \bfseries 0.783 & 0.663 +- 0.001 & 0.689 +- 0.001 & 0.613 \\
SIDER (26)                    & \bfseries 0.660 & 0.569 +- 0.030 & 0.574 +- 0.023 & 0.572 \\
Tox21 (12)                    & \bfseries 0.739 & 0.661 +- 0.025 & 0.652 +- 0.026 & 0.650 \\
BioSNAP (35)                  & \bfseries 0.797 & 0.664 +- 0.033 & 0.664 +- 0.017 & 0.443 \\
AVIDa-hIL6 (12)               & \bfseries 0.902 & 0.544 +- 0.035 & 0.755 +- 0.054 & 0.678 \\
\addlinespace[4pt]
\rowcolor{maintableblue!65}[0pt][\tabcolsep]
\multicolumn{5}{@{}l}{\textit{ROC-AUC (held out from molecular training)}} \\
FS-Mol held-out assays (157)  & \bfseries 0.760 & 0.640 +- 0.003 & 0.656 +- 0.002 & 0.616 \\
CYP450 (5)                    & \bfseries 0.850 & 0.598 +- 0.042 & 0.578 +- 0.029 & 0.601 \\
ToxCast (482)                 & \bfseries 0.632 & 0.574 +- 0.005 & 0.583 +- 0.007 & 0.563 \\
BindingDB (146)               & \bfseries 0.919 & 0.662 +- 0.013 & 0.731 +- 0.011 & 0.437 \\
AVIDa-hIL6 held-out (7)       & \bfseries 1.000 & 0.514 +- 0.081 & 0.715 +- 0.082 & 0.410 \\
\addlinespace[4pt]
\rowcolor{maintableblue!65}[0pt][\tabcolsep]
\multicolumn{5}{@{}l}{\textit{Multiclass accuracy}} \\
DrugBank DDI (86 classes)     & \bfseries 0.829 & 0.312 +- 0.012 & {---} & 0.309 \\
USPTO reactions (50 classes)  & \bfseries 0.930 & 0.779 +- 0.007 & {---} & 0.735 \\
\addlinespace[4pt]
\rowcolor{maintableblue!65}[0pt][\tabcolsep]
\multicolumn{5}{@{}l}{\textit{Multiclass macro-F1}} \\
DrugBank DDI (86 classes)     & \bfseries 0.758 & 0.031 +- 0.005 & {---} & 0.027 \\
USPTO reactions (50 classes)  & \bfseries 0.930 & 0.755 +- 0.006 & {---} & 0.712 \\
\bottomrule
\end{tabular*}
\end{table}

\subsection{Generalization to new tasks and new compounds}\label{sec:generalization}
Across eight groups held out from molecular training, MolReJev reaches a mean ROC-AUC of 0.814 using task descriptions and four retrieved examples at fixed weights (Table~\ref{tab:main}). Given four examples each, MolReJev leads Qwen3-4B in every held-out group, with the largest gains on held-out FS-Mol and ClinTox.

ProtoNet classifies a query by its distance to class prototypes formed by averaging the embeddings of labeled support molecules in each class \citep{snell2017prototypical}. With four retrieved molecules, MolReJev's ROC-AUC falls between the ProtoNet settings with 16 and 64 randomly sampled support molecules on the same 157 held-out FS-Mol assays (Table~\ref{tab:protonet}).
\begin{table}[!htbp]
\centering\small
\caption{ROC-AUC on the same 157 held-out FS-Mol assays. Bold and underline mark the highest and second-highest scores, respectively. Full-pool random forest results are in Table~\ref{tab:rf-full}.}
\label{tab:protonet}
\begin{tabular}{llr}
\toprule
\textbf{Method} & \textbf{Support / demos} & \textbf{ROC-AUC} \\
\midrule
ProtoNet & random 16 rows & 0.687 \\
ProtoNet & random 64 rows & \textbf{0.773} \\
Random forest & 16 rows & 0.604 \\
Random forest & 64 rows & 0.672 \\
kNN & most similar 4 & 0.666 \\
MolReJev & most similar 4 & \underline{0.760} \\
\bottomrule
\end{tabular}
\end{table}

\begin{table}[!t]
\centering\small
\caption{Comparison with molecular baselines. kNN and MolReJev use the same four references. RF uses 16 or 64 labels per task (full pools in Table~\ref{tab:rf-full}). MolecularGPT's multiclass entries use 1,000 queries per dataset. The other columns use the full test sets. On the MolecularGPT subsamples, MolReJev scores 0.829 accuracy and 0.758 macro-F1 on DDI and 0.930 on both metrics on USPTO. Bold marks the highest displayed score per row.}
\label{tab:classical}\label{tab:mgpt}
\sisetup{table-format=1.3, mode=text, reset-text-series=false}
\setlength{\tabcolsep}{5pt}
\begin{tabular*}{\linewidth}{@{\extracolsep{\fill}}l S S S S S@{}}
\toprule
\textbf{Dataset group} & \multicolumn{1}{c}{\textbf{kNN}} & \multicolumn{1}{c}{\textbf{RF 16}} & \multicolumn{1}{c}{\textbf{RF 64}} & \multicolumn{1}{c}{\textbf{MolecularGPT}} & \multicolumn{1}{c}{\cellcolor{maintableblue}\textbf{MolReJev}} \\
\midrule
\rowcolor{maintableblue!65}[0pt][\tabcolsep]
\multicolumn{6}{@{}l}{\textit{ROC-AUC (seen in molecular training)}} \\
FS-Mol (4507)                 & 0.664 & 0.628 & 0.693 & 0.695 & \bfseries 0.782 \\
BACE (1)                      & 0.841 & 0.473 & 0.679 & 0.760 & \bfseries 0.861 \\
BBBP (1)                      & 0.651 & 0.552 & 0.662 & \bfseries 0.718 & 0.694 \\
HIV (1)                       & 0.784 & 0.639 & 0.643 & 0.774 & \bfseries 0.821 \\
SIDER (27)                    & 0.638 & 0.524 & 0.526 & 0.559 & \bfseries 0.640 \\
Tox21 (12)                    & 0.658 & 0.592 & 0.606 & 0.664 & \bfseries 0.751 \\
BioSNAP (35)                  & 0.670 & 0.787 & \bfseries 0.800 & {---} & 0.797 \\
AVIDa-hIL6 (16)               & \bfseries 0.939 & 0.582 & 0.641 & {---} & 0.926 \\
\addlinespace[4pt]
\rowcolor{maintableblue!65}[0pt][\tabcolsep]
\multicolumn{6}{@{}l}{\textit{ROC-AUC (held out from molecular training)}} \\
FS-Mol held-out assays (157)  & 0.666 & 0.604 & 0.672 & 0.625 & \bfseries 0.760 \\
CYP450 (5)                    & 0.789 & 0.604 & 0.656 & \bfseries 0.866 & 0.834 \\
ToxCast (530)                 & 0.580 & 0.547 & 0.570 & 0.620 & \bfseries 0.628 \\
ClinTox (2)                   & 0.632 & 0.590 & 0.608 & 0.586 & \bfseries 0.810 \\
BindingDB (152)               & 0.864 & 0.850 & 0.913 & {---} & \bfseries 0.916 \\
AVIDa-hIL6 held-out (15)      & 0.943 & 0.493 & 0.687 & {---} & \bfseries 0.944 \\
AVIDa-SARS-CoV-2 (12)         & 0.688 & 0.434 & 0.515 & {---} & \bfseries 0.726 \\
AVIDa-hTNF$\alpha$ (1)        & \bfseries 0.902 & 0.738 & 0.720 & {---} & 0.893 \\
\midrule
Mean ROC-AUC (16 groups)      & 0.744 & 0.602 & 0.662 & {---} & \bfseries 0.799 \\
\addlinespace[4pt]
\rowcolor{maintableblue!65}[0pt][\tabcolsep]
\multicolumn{6}{@{}l}{\textit{Multiclass accuracy}} \\
DrugBank DDI (86 classes)     & 0.805 & {---} & {---} & 0.748 & \bfseries 0.843 \\
USPTO reactions (50 classes)  & 0.593 & {---} & {---} & 0.514 & \bfseries 0.912 \\
\addlinespace[4pt]
\rowcolor{maintableblue!65}[0pt][\tabcolsep]
\multicolumn{6}{@{}l}{\textit{Multiclass macro-F1}} \\
DrugBank DDI (86 classes)     & 0.736 & {---} & {---} & 0.698 & \bfseries 0.779 \\
USPTO reactions (50 classes)  & 0.586 & {---} & {---} & 0.514 & \bfseries 0.911 \\
\bottomrule
\end{tabular*}
\end{table}

\begin{figure}[!htbp]
\centering
\includegraphics[width=\linewidth]{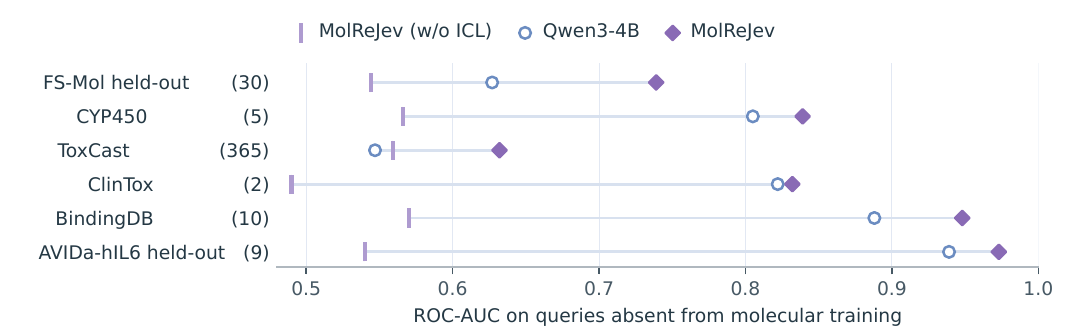}
\caption{ROC-AUC on query molecules absent from molecular training, averaged over qualifying held-out tasks (counts in parentheses). Qwen3-4B and MolReJev use the same four references. MolReJev (w/o ICL) uses the same trained weights with no references.}
\label{fig:unseen}
\end{figure}

\paragraph{Queries absent from training.} In held-out FS-Mol, 77.4\% of query molecules also occur in training records for other assays. We evaluate the queries with and without overlap separately (Appendix~\ref{app:overlap}). On compounds absent from training, four examples raise MolReJev's ROC-AUC from 0.544 to 0.739 across 30 qualifying FS-Mol assays. With four examples, MolReJev leads both its zero-example setting and Qwen3-4B with four examples in all six held-out groups with enough queries of both labels in each subset (Figure~\ref{fig:unseen}).

\subsection{Comparison with molecular baselines}\label{sec:predictors}
Table~\ref{tab:classical} compares MolReJev with kNN, task-specific random forests, and MolecularGPT. With access to the same four references as kNN, MolReJev leads in 14 of the 16 ROC-AUC groups and on both multiclass tasks. On USPTO, the accuracy gap is 0.319, showing that the same references can support markedly different predictions depending on how they are used. Similarity-weighted kNN leads on AVIDa-hIL6 and AVIDa-hTNF$\alpha$\footnote{\url{https://huggingface.co/datasets/COGNANO/AVIDa-hTNFa}}.

Access to labeled data changes the random forest comparison. MolReJev generally leads forests fitted to 16 or 64 labels per task, while full-pool forests lead in 11 of 16 ROC-AUC groups (Table~\ref{tab:rf-full}). MolReJev leads on both FS-Mol splits in either setting.

MolReJev leads MolecularGPT in eight of ten property prediction groups and on the matched DrugBank DDI and USPTO subsamples. MolecularGPT leads on BBBP and CYP450.

\subsection{Comparison with text generation}\label{sec:control}
Both trained systems start from Qwen3-4B and use the same training pool. On the same evaluation queries, both gain substantially from four examples (Table~\ref{tab:gen}). The difference in gains (MolReJev minus Qwen3-4B-Text) is $-0.020$ (95\% paired bootstrap CI $[-0.045, 0.006]$). With four examples, the ROC-AUC gap is $+0.004$ ($[-0.005, 0.013]$). Both intervals span zero. Appendix~\ref{app:generative} gives the bootstrap protocol and results by split.

\begin{table}[!htbp]
\centering\small
\caption{Effect of in-context examples for the two trained models. Mean ROC-AUC weights the 16 dataset groups equally. The gain is the change from zero to four examples.}
\label{tab:gen}
\begin{tabular}{@{}lrrr@{}}
\toprule
System & $k=0$ & $k=4$ & Gain from ICL \\
\midrule
MolReJev & 0.586 & 0.799 & $+0.213$ \\
Qwen3-4B-Text & 0.562 & 0.795 & $+0.233$ \\
\bottomrule
\end{tabular}
\end{table}

Qwen3-4B-Text generates a label string and provides a first-token ranking score for binary tasks. MolReJev returns a probability distribution over all candidates in one pass, including labels that share a prefix. For example, 66 of the 86 DrugBank labels begin with \texttt{\#Drug1}. MolReJev's predictions stay within the supplied label set. On USPTO, two generated answers fell outside that set (Appendix~\ref{app:cases}, Case~2).

Excluding retrieval, MolReJev processes 2.7--3.8 times as many records per second as Qwen3-4B-Text in our setup (Table~\ref{tab:speed}). The models use inputs of similar length. The largest speedup occurs on DrugBank DDI, where the text model decodes the longest label strings (Appendix~\ref{app:generative}).

\begin{table}[!htbp]
\centering\small
\caption{Inference throughput in records per second, excluding retrieval. Appendix~\ref{app:generative} describes each model's timing protocol. Bold marks the highest throughput in each row.}
\label{tab:speed}
\begin{tabular}{@{}lrrr@{}}
\toprule
Task & MolecularGPT & Qwen3-4B-Text & MolReJev \\
\midrule
Molecular groups & 84.85 & 38.70 & \textbf{120.40} \\
DrugBank DDI & 1.47 & 3.52 & \textbf{13.46} \\
USPTO & 2.37 & 11.93 & \textbf{32.09} \\
\bottomrule
\end{tabular}
\end{table}

\FloatBarrier
\section{Conclusion}
MolReJev combines Jev-style candidate scoring with in-context learning for molecular prediction. Retrieved examples improve predictions from the general-purpose Kev checkpoint and from the separately trained MolReJev. For the trained model, the gains extend to held-out tasks and compounds absent from training. MolReJev generally outperforms MolecularGPT on matched evaluations and nearest-neighbor prediction using the same examples. The paired comparison with Qwen3-4B-Text shows that both trained models benefit from in-context learning and attain similar mean ROC-AUC with four examples. MolReJev produces a probability distribution over the candidate labels in one forward pass and has higher measured throughput. A single decision model can thus use labeled examples at inference time to make predictions across molecular properties, drug--drug interactions, and reaction classes.

\bibliographystyle{iclr2025_conference}
\bibliography{references}

\begin{thebibliography}{47}
\providecommand{\natexlab}[1]{#1}
\providecommand{\url}[1]{\texttt{#1}}
\expandafter\ifx\csname urlstyle\endcsname\relax
  \providecommand{\doi}[1]{doi: #1}\else
  \providecommand{\doi}{doi: \begingroup \urlstyle{rm}\Url}\fi

\bibitem[Bai et~al.(2023)Bai, Miljkovi{\'c}, John, and Lu]{bai2023interpretable}
Peizhen Bai, Filip Miljkovi{\'c}, Bino John, and Haiping Lu.
\newblock Interpretable bilinear attention network with domain adaptation improves drug--target prediction.
\newblock \emph{Nature Machine Intelligence}, 5\penalty0 (2):\penalty0 126--136, 2023.

\bibitem[Bajusz et~al.(2015)Bajusz, R{\'a}cz, and H{\'e}berger]{bajusz2015tanimoto}
D{\'a}vid Bajusz, Anita R{\'a}cz, and K{\'a}roly H{\'e}berger.
\newblock Why is tanimoto index an appropriate choice for fingerprint-based similarity calculations?
\newblock \emph{Journal of cheminformatics}, 7\penalty0 (1):\penalty0 20, 2015.

\bibitem[Bemis \& Murcko(1996)Bemis and Murcko]{bemis1996properties}
Guy~W Bemis and Mark~A Murcko.
\newblock The properties of known drugs. 1. molecular frameworks.
\newblock \emph{Journal of medicinal chemistry}, 39\penalty0 (15):\penalty0 2887--2893, 1996.

\bibitem[Breiman(2001)]{breiman2001random}
Leo Breiman.
\newblock Random forests.
\newblock \emph{Machine learning}, 45\penalty0 (1):\penalty0 5--32, 2001.

\bibitem[Brown et~al.(2020)Brown, Mann, Ryder, Subbiah, Kaplan, Dhariwal, Neelakantan, Shyam, Sastry, Askell, et~al.]{brown2020language}
Tom Brown, Benjamin Mann, Nick Ryder, Melanie Subbiah, Jared~D Kaplan, Prafulla Dhariwal, Arvind Neelakantan, Pranav Shyam, Girish Sastry, Amanda Askell, et~al.
\newblock Language models are few-shot learners.
\newblock \emph{Advances in neural information processing systems}, 33:\penalty0 1877--1901, 2020.

\bibitem[Chen et~al.(2022)Chen, Zhong, Zha, Karypis, and He]{chen2022meta}
Yanda Chen, Ruiqi Zhong, Sheng Zha, George Karypis, and He~He.
\newblock Meta-learning via language model in-context tuning.
\newblock In \emph{Proceedings of the 60th Annual Meeting of the Association for Computational Linguistics (Volume 1: Long Papers)}, pp.\  719--730, 2022.

\bibitem[Dao(2024)]{dao2024flashattention}
Tri Dao.
\newblock Flashattention-2: Faster attention with better parallelism and work partitioning.
\newblock In \emph{International Conference on Learning Representations}, volume 2024, pp.\  35549--35562, 2024.

\bibitem[Deu{\ss}er et~al.(2026)Deu{\ss}er, Ramamurthy, Sparrenberg, and Sifa]{deusser2026evaluating}
Tobias Deu{\ss}er, Rajkumar Ramamurthy, Lorenz Sparrenberg, and Rafet Sifa.
\newblock Evaluating and benchmarking the system one model jev.
\newblock \emph{arXiv preprint arXiv:2609.37647}, 2026.

\bibitem[Du et~al.(2026)Du, Zhao, Zhao, Li, Gao, Wang, Xu, Cui, and Qin]{du2026reliable}
Yanrui Du, Danyang Zhao, Sendong Zhao, Wenhao Li, Yibo Gao, Haochun Wang, Zhuoer Xu, Shiwen Cui, and Bing Qin.
\newblock How reliable are system-one decisions in medicine? a comparative study of jev, jev-like models, and llms.
\newblock \emph{arXiv preprint arXiv:2610.18529}, 2026.

\bibitem[Durant et~al.(2002)Durant, Leland, Henry, and Nourse]{durant2002reoptimization}
Joseph~L Durant, Burton~A Leland, Douglas~R Henry, and James~G Nourse.
\newblock Reoptimization of mdl keys for use in drug discovery.
\newblock \emph{Journal of chemical information and computer sciences}, 42\penalty0 (6):\penalty0 1273--1280, 2002.

\bibitem[Efron \& Tibshirani(1993)Efron and Tibshirani]{efron1993bootstrap}
Bradley Efron and Robert~J. Tibshirani.
\newblock \emph{An Introduction to the Bootstrap}.
\newblock Chapman and Hall, 1993.
\newblock \doi{10.1201/9780429246593}.

\bibitem[Hanley \& McNeil(1982)Hanley and McNeil]{hanley1982meaning}
James~A Hanley and Barbara~J McNeil.
\newblock The meaning and use of the area under a receiver operating characteristic (roc) curve.
\newblock \emph{Radiology}, 143\penalty0 (1):\penalty0 29--36, 1982.

\bibitem[Hu et~al.(2021)Hu, Shen, Wallis, Allen-Zhu, Li, Wang, Wang, and Chen]{hu2021lora}
Edward~J Hu, Yelong Shen, Phillip Wallis, Zeyuan Allen-Zhu, Yuanzhi Li, Shean Wang, Lu~Wang, and Weizhu Chen.
\newblock Lora: Low-rank adaptation of large language models.
\newblock \emph{arXiv preprint arXiv:2106.09685}, 2021.

\bibitem[Huang et~al.(2021)Huang, Fu, Gao, Zhao, Roohani, Leskovec, Coley, Xiao, Sun, and Zitnik]{huang2021therapeutics}
Kexin Huang, Tianfan Fu, Wenhao Gao, Yue Zhao, Yusuf Roohani, Jure Leskovec, Connor~W Coley, Cao Xiao, Jimeng Sun, and Marinka Zitnik.
\newblock Therapeutics data commons: Machine learning datasets and tasks for drug discovery and development.
\newblock \emph{arXiv preprint arXiv:2102.09548}, 2021.

\bibitem[Ji et~al.(2023)Ji, Zhang, Wu, Wu, Li, Huang, Xu, Rong, Ren, Xue, et~al.]{ji2023drugood}
Yuanfeng Ji, Lu~Zhang, Jiaxiang Wu, Bingzhe Wu, Lanqing Li, Long-Kai Huang, Tingyang Xu, Yu~Rong, Jie Ren, Ding Xue, et~al.
\newblock Drugood: Out-of-distribution dataset curator and benchmark for ai-aided drug discovery--a focus on affinity prediction problems with noise annotations.
\newblock In \emph{Proceedings of the AAAI Conference on Artificial Intelligence}, volume~37, pp.\  8023--8031, 2023.

\bibitem[Landrum et~al.(2013)]{landrum2013rdkit}
Greg Landrum et~al.
\newblock Rdkit documentation.
\newblock \emph{Release}, 1\penalty0 (1-79):\penalty0 4, 2013.

\bibitem[Leskovec()]{leskovecstanford}
J~Leskovec.
\newblock Stanford biomedical network dataset collection.

\bibitem[Li et~al.(2024)Li, Liu, Fan, Wei, Liu, Tang, and Li]{li2024empowering}
Jiatong Li, Yunqing Liu, Wenqi Fan, Xiao-Yong Wei, Hui Liu, Jiliang Tang, and Qing Li.
\newblock Empowering molecule discovery for molecule-caption translation with large language models: A chatgpt perspective.
\newblock \emph{IEEE transactions on knowledge and data engineering}, 36\penalty0 (11):\penalty0 6071--6083, 2024.

\bibitem[Li et~al.(2025)Li, Liu, Ding, Fan, Li, and Li]{li2025large}
Jiatong Li, Wei Liu, Zhihao Ding, Wenqi Fan, Yuqiang Li, and Qing Li.
\newblock Large language models are in-context molecule learners.
\newblock \emph{IEEE Transactions on Knowledge and Data Engineering}, 37\penalty0 (7):\penalty0 4131--4143, 2025.

\bibitem[Li et~al.(2026{\natexlab{a}})Li, Liu, Liu, Lei, Zhang, Fan, Zhou, Li, and Li]{li2026molreflect}
Jiatong Li, Yunqing Liu, Wei Liu, Jingdi Lei, Di~Zhang, Wenqi Fan, Dongzhan Zhou, Yuqiang Li, and Qing Li.
\newblock Molreflect: Towards in-context fine-grained alignments between molecules and texts.
\newblock \emph{IEEE Transactions on Knowledge and Data Engineering}, 2026{\natexlab{a}}.

\bibitem[Li et~al.(2026{\natexlab{b}})Li, Ren, Wang, Wei, and Bian]{li2026chemical}
Jiatong Li, Yuxuan Ren, Weida Wang, Xiaoyong Wei, and Yatao Bian.
\newblock Chemical chain-of-thought functions as a hallucination-prone molecular scratchpad.
\newblock \emph{arXiv preprint arXiv:2607.20935}, 2026{\natexlab{b}}.

\bibitem[Li et~al.(2026{\natexlab{c}})Li, Wang, Zheng, Zhang, Bian, Wei, and Li]{li2026llms}
Jiatong Li, Weida Wang, Changmeng Zheng, Shufei Zhang, Yatao Bian, Xiao-yong Wei, and Qing Li.
\newblock Do llms truly generalize in the molecular domain? a perturbation-based analysis.
\newblock \emph{arXiv preprint arXiv:2607.01800}, 2026{\natexlab{c}}.

\bibitem[Liu et~al.(2022)Liu, Shen, Zhang, Dolan, Carin, and Chen]{liu2022makes}
Jiachang Liu, Dinghan Shen, Yizhe Zhang, William~B Dolan, Lawrence Carin, and Weizhu Chen.
\newblock What makes good in-context examples for gpt-3?
\newblock In \emph{Proceedings of Deep Learning Inside Out (DeeLIO 2022): The 3rd workshop on knowledge extraction and integration for deep learning architectures}, pp.\  100--114, 2022.

\bibitem[Liu et~al.(2007)Liu, Lin, Wen, Jorissen, and Gilson]{liu2007bindingdb}
Tiqing Liu, Yuhmei Lin, Xin Wen, Robert~N Jorissen, and Michael~K Gilson.
\newblock Bindingdb: a web-accessible database of experimentally determined protein--ligand binding affinities.
\newblock \emph{Nucleic acids research}, 35\penalty0 (suppl\_1):\penalty0 D198--D201, 2007.

\bibitem[Liu et~al.(2024)Liu, Ding, Zhou, Fan, and Tan]{liu2024moleculargpt}
Yuyan Liu, Sirui Ding, Sheng Zhou, Wenqi Fan, and Qiaoyu Tan.
\newblock Moleculargpt: Open large language model (llm) for few-shot molecular property prediction.
\newblock \emph{arXiv preprint arXiv:2406.12950}, 2024.

\bibitem[Loshchilov \& Hutter(2017)Loshchilov and Hutter]{loshchilov2017decoupled}
Ilya Loshchilov and Frank Hutter.
\newblock Decoupled weight decay regularization.
\newblock \emph{arXiv preprint arXiv:1711.05101}, 2017.

\bibitem[Lu et~al.(2026)Lu, Li, and Bian]{lu2026reasoning}
Kaiyuan Lu, Jiatong Li, and Yatao Bian.
\newblock Reasoning or just prompting? probing the mechanism of latent chemical reasoning.
\newblock \emph{arXiv preprint arXiv:2610.34173}, 2026.

\bibitem[Lyu \& Xia(2026)Lyu and Xia]{lyu2026decision}
Hanjia Lyu and Yinglong Xia.
\newblock Decision-oriented recommendation reranking: An empirical study of its promise and limitations.
\newblock \emph{arXiv preprint arXiv:2609.40241}, 2026.

\bibitem[Mendez et~al.(2019)Mendez, Gaulton, Bento, Chambers, De~Veij, F{\'e}lix, Magari{\~n}os, Mosquera, Mutowo, Nowotka, et~al.]{mendez2019chembl}
David Mendez, Anna Gaulton, A~Patr{\'\i}cia Bento, Jon Chambers, Marleen De~Veij, Eloy F{\'e}lix, Mar{\'\i}a~Paula Magari{\~n}os, Juan~F Mosquera, Prudence Mutowo, Micha{\l} Nowotka, et~al.
\newblock Chembl: towards direct deposition of bioassay data.
\newblock \emph{Nucleic acids research}, 47\penalty0 (D1):\penalty0 D930--D940, 2019.

\bibitem[Min et~al.(2022)Min, Lewis, Zettlemoyer, and Hajishirzi]{min2022metaicl}
Sewon Min, Mike Lewis, Luke Zettlemoyer, and Hannaneh Hajishirzi.
\newblock Metaicl: Learning to learn in context.
\newblock In \emph{Proceedings of the 2022 conference of the North American chapter of the Association for Computational Linguistics: Human Language Technologies}, pp.\  2791--2809, 2022.

\bibitem[Phan et~al.(2026)Phan, Nguyen~Song, and Stadler]{phan2026synrxn}
Tieu-Long Phan, Nhu-Ngoc Nguyen~Song, and Peter~F Stadler.
\newblock Synrxn: An open benchmark and curated dataset for computational reaction modeling.
\newblock \emph{Scientific Data}, 13\penalty0 (1):\penalty0 625, 2026.

\bibitem[Rogers \& Hahn(2010)Rogers and Hahn]{rogers2010extended}
David Rogers and Mathew Hahn.
\newblock Extended-connectivity fingerprints.
\newblock \emph{Journal of chemical information and modeling}, 50\penalty0 (5):\penalty0 742--754, 2010.

\bibitem[Rubin et~al.(2022)Rubin, Herzig, and Berant]{rubin2022learning}
Ohad Rubin, Jonathan Herzig, and Jonathan Berant.
\newblock Learning to retrieve prompts for in-context learning.
\newblock In \emph{Proceedings of the 2022 conference of the North American chapter of the association for computational linguistics: human language technologies}, pp.\  2655--2671, 2022.

\bibitem[Sanh et~al.(2021)Sanh, Webson, Raffel, Bach, Sutawika, Alyafeai, Chaffin, Stiegler, Scao, Raja, et~al.]{sanh2021multitask}
Victor Sanh, Albert Webson, Colin Raffel, Stephen~H Bach, Lintang Sutawika, Zaid Alyafeai, Antoine Chaffin, Arnaud Stiegler, Teven~Le Scao, Arun Raja, et~al.
\newblock Multitask prompted training enables zero-shot task generalization.
\newblock \emph{arXiv preprint arXiv:2110.08207}, 2021.

\bibitem[Schneider et~al.(2015)Schneider, Lowe, Sayle, and Landrum]{schneider2015development}
Nadine Schneider, Daniel~M Lowe, Roger~A Sayle, and Gregory~A Landrum.
\newblock Development of a novel fingerprint for chemical reactions and its application to large-scale reaction classification and similarity.
\newblock \emph{Journal of chemical information and modeling}, 55\penalty0 (1):\penalty0 39--53, 2015.

\bibitem[Snell et~al.(2017)Snell, Swersky, and Zemel]{snell2017prototypical}
Jake Snell, Kevin Swersky, and Richard Zemel.
\newblock Prototypical networks for few-shot learning.
\newblock \emph{Advances in neural information processing systems}, 30, 2017.

\bibitem[Stanley et~al.(2021)Stanley, Bronskill, Maziarz, Misztela, Lanini, Segler, Schneider, and Brockschmidt]{stanley2021fs}
Megan Stanley, John~F Bronskill, Krzysztof Maziarz, Hubert Misztela, Jessica Lanini, Marwin Segler, Nadine Schneider, and Marc Brockschmidt.
\newblock Fs-mol: A few-shot learning dataset of molecules.
\newblock In \emph{Thirty-fifth Conference on Neural Information Processing Systems Datasets and Benchmarks Track (Round 2)}, 2021.

\bibitem[Tsuruta et~al.(2023)Tsuruta, Yamazaki, Maeda, Tamura, Wei, Mariet, Phloyphisut, Shimokawa, Ledsam, Colwell, et~al.]{tsuruta2023avida}
Hirofumi Tsuruta, Hiroyuki Yamazaki, Ryota Maeda, Ryotaro Tamura, Jennifer Wei, Zelda~E Mariet, Poomarin Phloyphisut, Hidetoshi Shimokawa, Joseph~R Ledsam, Lucy Colwell, et~al.
\newblock Avida-hil6: a large-scale vhh dataset produced from an immunized alpaca for predicting antigen-antibody interactions.
\newblock \emph{Advances in Neural Information Processing Systems}, 36:\penalty0 42077--42096, 2023.

\bibitem[Tsuruta et~al.(2024)Tsuruta, Yamazaki, Maeda, Tamura, and Imura]{tsuruta2024sars}
Hirofumi Tsuruta, Hiroyuki Yamazaki, Ryota Maeda, Ryotaro Tamura, and Akihiro Imura.
\newblock A sars-cov-2 interaction dataset and vhh sequence corpus for antibody language models.
\newblock \emph{Advances in Neural Information Processing Systems}, 37:\penalty0 116149--116171, 2024.

\bibitem[{TypeSafe AI}(2026)]{almeida2026jev}
{TypeSafe AI}.
\newblock {Introducing System One Models \& Jev}.
\newblock \url{https://typesafe.ai/blog/introducing-system-one-models-and-jev}, September 2026.

\bibitem[Vinyals et~al.(2015)Vinyals, Fortunato, and Jaitly]{vinyals2015pointer}
Oriol Vinyals, Meire Fortunato, and Navdeep Jaitly.
\newblock Pointer networks.
\newblock \emph{Advances in neural information processing systems}, 28, 2015.

\bibitem[Wang et~al.(2021)Wang, Abuduweili, Yao, and Dou]{wang2021property}
Yaqing Wang, Abulikemu Abuduweili, Quanming Yao, and Dejing Dou.
\newblock Property-aware relation networks for few-shot molecular property prediction.
\newblock \emph{Advances in Neural Information Processing Systems}, 34:\penalty0 17441--17454, 2021.

\bibitem[Wei et~al.(2021)Wei, Bosma, Zhao, Guu, Yu, Lester, Du, Dai, and Le]{wei2021finetuned}
Jason Wei, Maarten Bosma, Vincent~Y Zhao, Kelvin Guu, Adams~Wei Yu, Brian Lester, Nan Du, Andrew~M Dai, and Quoc~V Le.
\newblock Finetuned language models are zero-shot learners.
\newblock \emph{arXiv preprint arXiv:2109.01652}, 2021.

\bibitem[Weininger(1988)]{weininger1988smiles}
David Weininger.
\newblock Smiles, a chemical language and information system. 1. introduction to methodology and encoding rules.
\newblock \emph{Journal of chemical information and computer sciences}, 28\penalty0 (1):\penalty0 31--36, 1988.

\bibitem[Wishart et~al.(2018)Wishart, Feunang, Guo, Lo, Marcu, Grant, Sajed, Johnson, Li, Sayeeda, et~al.]{wishart2018drugbank}
David~S Wishart, Yannick~D Feunang, An~C Guo, Elvis~J Lo, Ana Marcu, Jason~R Grant, Tanvir Sajed, Daniel Johnson, Carin Li, Zinat Sayeeda, et~al.
\newblock Drugbank 5.0: a major update to the drugbank database for 2018.
\newblock \emph{Nucleic acids research}, 46\penalty0 (D1):\penalty0 D1074--D1082, 2018.

\bibitem[Wu et~al.(2018)Wu, Ramsundar, Feinberg, Gomes, Geniesse, Pappu, Leswing, and Pande]{wu2018moleculenet}
Zhenqin Wu, Bharath Ramsundar, Evan~N Feinberg, Joseph Gomes, Caleb Geniesse, Aneesh~S Pappu, Karl Leswing, and Vijay Pande.
\newblock Moleculenet: a benchmark for molecular machine learning.
\newblock \emph{Chemical science}, 9\penalty0 (2):\penalty0 513--530, 2018.

\bibitem[Yang et~al.(2025)Yang, Li, Yang, Zhang, Hui, Zheng, Yu, Gao, Huang, Lv, et~al.]{yang2025qwen3}
An~Yang, Anfeng Li, Baosong Yang, Beichen Zhang, Binyuan Hui, Bo~Zheng, Bowen Yu, Chang Gao, Chengen Huang, Chenxu Lv, et~al.
\newblock Qwen3 technical report.
\newblock \emph{arXiv preprint arXiv:2505.09388}, 2025.

\end{thebibliography}
\appendix
\section{Experimental Protocol}\label{app:details}

\subsection{Datasets and splits}

\begin{table}[!htbp]
\centering\small
\caption{Datasets. ``Tasks'' counts tasks with evaluation queries after filtering. ``Train records'' gives the number of training records constructed from pool rows; 0 indicates an evaluation-only dataset. DrugBank DDI and USPTO each have 1,000 validation queries, excluded from this table.}
\label{tab:data}
\begin{adjustbox}{max width=\linewidth}
\begin{tabular}{@{}llrrr@{}}
\toprule
 & Dataset & Tasks & Train records & Eval queries \\
\midrule
\multirow{8}{*}{\rotatebox{90}{Used in training}}
 & FS-Mol (train assays) & 4,507 & 379,912 & 211,145 \\
 & BACE & 1 & 1,024 & 152 \\
 & BBBP & 1 & 1,024 & 197 \\
 & HIV & 1 & 1,024 & 256 \\
 & SIDER (27 organ classes) & 27 & 27,648 & 3,861 \\
 & Tox21 (12 assays) & 12 & 12,288 & 3,072 \\
 & BioSNAP & 35 & 812 & 309 \\
 & AVIDa-hIL6 & 16 & 16,384 & 4,096 \\
\addlinespace
\multicolumn{5}{@{}l}{\textit{Multi-class datasets}} \\
 & DrugBank DDI (86 classes) & 1 & 40,000 & 38,362 \\
 & USPTO reactions (50 classes) & 1 & 9,000 & 10,000 \\
\midrule
\multirow{8}{*}{\rotatebox{90}{Not used in training}}
 & FS-Mol held-out assays & 157 & 0 & 11,676 \\
 & CYP450 & 5 & 0 & 1,280 \\
 & ToxCast & 530 & 0 & 83,342 \\
 & ClinTox & 2 & 0 & 294 \\
 & BindingDB & 152 & 0 & 4,996 \\
 & \shortstack[l]{AVIDa-hIL6\\held-out} & 15 & 0 & 3,840 \\
 & AVIDa-SARS-CoV-2 & 12 & 0 & 3,072 \\
 & AVIDa-hTNF$\alpha$ & 1 & 0 & 256 \\
\midrule
 & Total (18 groups) & & 489,116 & 380,206 \\
\bottomrule
\end{tabular}
\end{adjustbox}
\end{table}

For the binary molecular tasks, we split the labeled records into a reference pool and separate evaluation queries. All demonstrations come from the pool.
We retain up to 256 queries per task, preserving the class ratio.
We exclude tasks with fewer than three queries of either class or fewer than five pool records.
In training, we draw queries from the pool, excluding each query from its own demonstrations. We cap training queries at 128 per FS-Mol assay and 1,024 per task elsewhere.

\emph{FS-Mol} \citep{stanley2021fs}. We use the official train/test assay split. We require at least 60 records, both activity labels, and a ChEMBL assay description, retaining 4,624 of the 26,774 training assays in our data files. Within each assay, 25\,\% of each class is held out for queries. The query and pool size requirements leave 4,507 training assays and all 157 held-out assays.
\emph{BACE, BBBP, HIV, SIDER, Tox21, ToxCast, ClinTox} \citep{wu2018moleculenet}: we split on Bemis--Murcko scaffolds \citep{bemis1996properties} in an 8:1:1 ratio. Train and validation molecules form the pool, and test molecules are queries.
\emph{BioSNAP} (\citeauthor{leskovecstanford}) and \emph{BindingDB} \citep{liu2007bindingdb}: DrugBAN's published split \citep{bai2023interpretable}; train and validation rows are the pool, test rows are queries; one task per target protein. BioSNAP retains 35 of 1,822 targets and BindingDB retains 152 of 1,441 targets after size filtering.
\emph{AVIDa-hIL6} \citep{tsuruta2023avida}: one task per antigen variant; the 15 mutants listed in the dataset paper are held out as whole tasks (evaluation-only); wild type and the other 15 mutants are training tasks, each with a 25/75 query/pool split.
\emph{CYP450:} the file's own train/test split; the training part is the pool, the test part is queries; evaluation-only.
\emph{AVIDa-SARS-CoV-2} \citep{tsuruta2024sars}: published train/test files; train rows are the pool, test rows are queries; evaluation-only.
\emph{AVIDa-hTNF$\alpha$}: single antigen; 25\,\% of each class as queries; evaluation-only.

\paragraph{DrugBank DDI.} the TDC \citep{huang2021therapeutics} DrugBank records \citep{wishart2018drugbank}, shuffled with seed 42 into 70\% training, 10\% validation and 20\% test. Training queries are 40,000 records drawn from the training split with class stratification, taking at least $\min(N_c, 50)$ records of each class and the remainder in proportion. Validation queries are 1,000 records drawn at random from the validation split. The evaluation queries are the whole test split, 38,362 records. Demonstrations come from all 134,265 rows of the training split, a larger set than the training queries.

\paragraph{USPTO.} the SynRXN \texttt{schneider\_b} split \citep{phan2026synrxn,schneider2015development}, whose training part holds 200 reactions of each of the 50 classes. We hold out 20 of each class as validation queries, 1,000 records excluded from the reference pool, and use the remaining 9,000 as both the training queries and the reference pool. The evaluation queries are 200 reactions of each class sampled at random from the official test set, 10,000 of its 40,000 records, so the evaluation set is balanced across classes.

\medskip
\noindent\begin{minipage}{\linewidth}
\subsection{Model configurations}

{\centering\small
\providecommand{\cmark}{\ensuremath{\checkmark}}
\providecommand{\xmark}{\ensuremath{\times}}
\captionof{table}{Model configurations on the Qwen3-4B backbone. Molecular training uses rank-16 LoRA adapters. CE denotes cross-entropy.}
\label{tab:configs}
\begin{adjustbox}{max width=\linewidth}
\begin{tabular}{@{}lccc@{}}
\toprule
 & Qwen3-4B & Qwen3-4B-Text & MolReJev \\
\midrule
Starting checkpoint & Qwen3-4B & Qwen3-4B & Qwen3-4B \\
Prediction head & LM head & LM head & Pointer head \\
Molecular training & \xmark & \cmark & \cmark \\
Molecular training loss & -- & Vocabulary CE & Candidate CE \\
Peak learning rate & -- & $5\times10^{-5}$ & $5\times10^{-5}$ \\
Maximum record length & -- & 4,096 & 3,072 \\
\bottomrule
\end{tabular}
\end{adjustbox}
\par}
\end{minipage}\par

\subsection{Retrieval and prompt construction}

Table~\ref{tab:retrieval-protocol} specifies the similarity measure and reference pool for each input type. MACCS fingerprints are computed with RDKit \citep{landrum2013rdkit}.
\begin{table}[!htbp]
\centering
\caption{Demonstration retrieval. Each query type uses the representation and
similarity listed here; the four most similar references are placed in the
prompt in decreasing order of similarity, with ties broken by row order. A
training query never retrieves itself.}
\label{tab:retrieval-protocol}
\small
\begin{adjustbox}{max width=\linewidth}
\begin{tabular}{@{}l l l l@{}}
\toprule
Query type & Representation & Similarity & Reference pool \\
\midrule
Small molecule & SMILES & MACCS Tanimoto & the task's pool \\
Antibody & amino acid sequence & cosine, TF--IDF of trigrams & the task's pool \\
Drug pair & two SMILES, order kept & mean of the two MACCS & DrugBank training \\
 & & Tanimoto scores, drug 1 to & split (first 70\% \\
 & & drug 1 and drug 2 to drug 2 & after shuffling with seed 42) \\
Reaction & reactants \texttt{>>} products & cosine, TF--IDF of character & official training \\
 & & trigrams & split less the \\
 & & & validation rows \\
\bottomrule
\end{tabular}
\end{adjustbox}
\end{table}

For USPTO, the TF--IDF vocabulary is fitted only on the reference pool, with a minimum term frequency of two.

\paragraph{Prompt templates.}
The state begins with the task description (minus its final sentence), followed by any context sequence (antigen or target protein), then ``Here are some examples.'', then $k$ demonstration pairs (each a SMILES or antibody sequence with a task label), and finally the query in the same format without a label.
The question is the final sentence of the task description, used verbatim.

For the decision model, the state and question are packed into a single sequence using Kev's delimiter tokens:

\begin{verbatim}
<state> [state text]
<q> [question] <opt> no </opt> <opt> yes </opt> <decide>
\end{verbatim}

\noindent The delimiters reuse existing Qwen special tokens (\texttt{<state>}~=~\texttt{<|fim\_prefix|>}, \texttt{<q>}~=~\texttt{<|fim\_middle|>}, \texttt{<opt>}~=~\texttt{<|box\_start|>}, \texttt{</opt>}~=~\texttt{<|box\_end|>}, \texttt{<decide>}~=~\texttt{<|fim\_suffix|>}).
The pointer head reads the final hidden states at the \texttt{</opt>} tokens and at \texttt{<decide>} to compute scores as described in Section~\ref{sec:scoring}.

For Qwen3-4B-Text, we concatenate the state, question, and candidate labels listed one per line, followed by the JSON answer prefix \texttt{\{"answer": "}. The generated answer text must exactly match an option string; invalid outputs count as errors.

\noindent\begin{minipage}{\linewidth}
The state for the PLK3 query in Figures~\ref{fig:intro} and~\ref{fig:method} is shown below, with line breaks added to fit.

\begin{verbatim}
PUBCHEM_BIOASSAY: Navigating the Kinome.
(Class of assay: other) Panel member name: PLK3.

Here are some examples.
SMILES: COc1nccn2c(-c3ccnc(NCC(C)(C)CO)n3)c(-c3ccc(F)cc3F)nc12
label: Yes
SMILES: CC(C)COc1nccn2c(-c3ccnc(NCC(C)(C)CO)n3)c(-c3ccc(F)cc3)nc12
label: No
SMILES: CC(C)Oc1nccn2c(-c3ccnc(NCC(C)(C)CO)n3)c(-c3ccc(F)cc3F)nc12
label: No
SMILES: CNc1nccn2c(-c3ccnc(NCC(C)(C)CO)n3)c(-c3ccc(F)cc3)nc12
label: No
SMILES: COc1nccn2c(-c3ccnc(NCC(C)(C)CO)n3)c(-c3ccc(F)cc3)nc12
\end{verbatim}

\noindent Question: \texttt{Is this molecule effective to this assay?}
\end{minipage}

For antibody tasks (AVIDa), the field name is \texttt{Sequence} instead of \texttt{SMILES}, and the antigen sequence appears before the demonstrations.
For BioSNAP and BindingDB, the target protein sequence is included in the same position.
Task descriptions come from released MolecularGPT prompts \citep{liu2024moleculargpt} (BACE, BBBP, HIV, Tox21, ToxCast, CYP450), ChEMBL assay descriptions \citep{mendez2019chembl} (FS-Mol), or templates we wrote (SIDER, ClinTox, BioSNAP, BindingDB, AVIDa).

\subsection{Evaluation and baseline protocols}

\paragraph{Evaluation.}
We assign tied scores their average rank and omit tasks with only one label from the ROC-AUC mean. Macro-F1 averages over classes present in the evaluation labels. Table~\ref{tab:data} gives task and query counts for the main evaluation. In analyses limited to the first 32 queries, some tasks no longer contain both labels. The demonstration-count and retrieval tables report the number of tasks contributing to each mean.

\paragraph{Ablation protocol.}
For the demonstration-count sweep, the zero- and four-example conditions use the same query IDs. The one- and two-example prompts use the nearest members of the stored four; for eight examples we retrieve again from the task pool. The retrieval comparison fixes the count at four and varies the choice of examples: most similar, randomly sampled, class-balanced random, or least similar.

\paragraph{MolecularGPT.}
We use the authors' LoRA adapter on Llama-2-7B-chat and format the input with their prompt template. For binary tasks, we score queries by the Yes--No logit difference. The comparison covers tasks that require neither protein nor antibody sequence input.

\paragraph{Random forest and kNN.}
Random forests \citep{breiman2001random} use 100 trees with ECFP4 count features \citep{rogers2010extended} for small molecules or hashed amino acid trigram counts for antibodies. We sample 16 or 64 pool records with class stratification, using the whole pool if it is smaller. The full-pool forest fits all pool records. For DDI, we concatenate the two drugs' Morgan count vectors in their given order. For USPTO, we subtract the summed reactant fingerprints from the summed product fingerprints. The kNN predictor uses the same four neighbors as MolReJev. Ties in multi-class voting are resolved in favour of the label of the nearest tied neighbour. Full-pool results and pool sizes are reported in Table~\ref{tab:rf-full}.

\paragraph{ProtoNet.}
We use the FS-Mol checkpoint for ProtoNet \citep{snell2017prototypical,stanley2021fs}, a metric-based predictor trained in episodes on molecular graphs and fingerprints. For each of the 157 held-out FS-Mol assays, its support comprises 16 or 64 randomly sampled pool molecules. Table~\ref{tab:protonet} compares these settings with MolReJev using four retrieved examples on the same queries.

\Needspace{8\baselineskip}
\section{Additional Baseline Results}

\subsection{Text generation}\label{app:generative}

Qwen3-4B-Text and MolReJev share a backbone, a training pool, and the optimization settings described in Section~\ref{sec:training}. Qwen3-4B-Text generates a JSON answer. Table~\ref{tab:trained-text} gives the per-group results, including the zero-example condition for the trained text baseline.

\begin{table}[!htbp]
\centering\small
\caption{Per-group results for the text and decision models. Qwen3-4B (text prompt) uses the same base checkpoint as Qwen3-4B in Table~\ref{tab:main} and the prompt format of Qwen3-4B-Text. Binary groups report ROC-AUC, and multiclass groups report accuracy / macro-F1. The overall mean weights the 16 ROC-AUC groups equally.}
\label{tab:base-text}
\label{tab:trained-text}
\begin{tabular}{@{}lcccc@{}}
\toprule
 & & \multicolumn{2}{c}{Qwen3-4B-Text} & \\
\cmidrule(lr){3-4}
Dataset (tasks) & \shortstack{Qwen3-4B\\(text prompt, $k=4$)} & $k=0$ & $k=4$ & MolReJev ($k=4$) \\
\midrule
\multicolumn{5}{@{}l}{\textit{Datasets used in training}} \\
FS-Mol (4507) & 0.634 & 0.627 & 0.778 & 0.782 \\
BACE (1) & 0.795 & 0.595 & 0.859 & 0.861 \\
BBBP (1) & 0.675 & 0.629 & 0.705 & 0.694 \\
HIV (1) & 0.778 & 0.749 & 0.795 & 0.821 \\
SIDER (27) & 0.609 & 0.552 & 0.669 & 0.640 \\
Tox21 (12) & 0.633 & 0.588 & 0.730 & 0.751 \\
BioSNAP (35) & 0.631 & 0.525 & 0.833 & 0.797 \\
AVIDa-hIL6 (16) & 0.933 & 0.613 & 0.942 & 0.926 \\
\addlinespace
\multicolumn{5}{@{}l}{\textit{Multi-class datasets (accuracy / macro-F1)}} \\
DrugBank DDI (86 classes) & 0.616 / 0.117 & --- & 0.839 / 0.782 & 0.843 / 0.779 \\
USPTO reactions (50 classes) & 0.558 / 0.551 & --- & 0.935 / 0.934 & 0.912 / 0.911 \\
\addlinespace
\multicolumn{5}{@{}l}{\textit{Datasets not used in training}} \\
FS-Mol held-out (157) & 0.636 & 0.628 & 0.760 & 0.760 \\
CYP450 (5) & 0.800 & 0.657 & 0.826 & 0.834 \\
ToxCast (530) & 0.566 & 0.546 & 0.624 & 0.628 \\
ClinTox (2) & 0.683 & 0.373 & 0.723 & 0.810 \\
BindingDB (152) & 0.840 & 0.509 & 0.914 & 0.916 \\
AVIDa-hIL6 held-out (15) & 0.934 & 0.604 & 0.942 & 0.944 \\
AVIDa-SARS-CoV-2 (12) & 0.681 & 0.510 & 0.719 & 0.726 \\
AVIDa-hTNF$\alpha$ (1) & 0.867 & 0.283 & 0.903 & 0.893 \\
\midrule
Mean, 16 ROC-AUC groups & 0.731 & 0.562 & 0.795 & 0.799 \\
\bottomrule
\end{tabular}
\end{table}

With four demonstrations, the trained text baseline produces zero invalid outputs among 38,362 DrugBank DDI queries and two among 10,000 USPTO queries, giving valid-option rates of 100\% and 99.98\%, respectively.

\paragraph{Paired comparison across example counts.}
For each task, we compute the change in ROC-AUC from $k=0$ to $k=4$ for each model and subtract the text model's gain from MolReJev's. We average within each group and weight the groups equally. The 95\% bootstrap percentile intervals \citep{efron1993bootstrap} use the 2.5th and 97.5th percentiles of 2,000 draws with seed~0. We resample tasks with replacement within each group. In groups with fewer than 30 tasks, we also resample queries with replacement within each selected task. The four conditions share the same draw to preserve pairing. Table~\ref{tab:paired-icl} reports the differences.

\begin{table}[!htbp]
\centering\small
\caption{Paired ROC-AUC differences (MolReJev minus Qwen3-4B-Text). The gain difference compares the changes from zero to four examples. The held-out gain interval ends at approximately $-0.00025$ before rounding.}
\label{tab:paired-icl}
\begin{tabular}{@{}llrc@{}}
\toprule
Comparison & Dataset groups & Difference & 95\% CI \\
\midrule
Gain difference & All (16) & $-0.020$ & $[-0.045, 0.006]$ \\
Gain difference & Seen (8) & $-0.001$ & $[-0.032, 0.030]$ \\
Gain difference & Held out (8) & $-0.039$ & $[-0.077, 0.000]$ \\
Four examples & All (16) & $+0.004$ & $[-0.005, 0.013]$ \\
Zero examples & Held out (8) & $+0.052$ & $[0.012, 0.089]$ \\
\bottomrule
\end{tabular}
\end{table}

The zero-example difference on held-out groups is concentrated in AVIDa-hTNF$\alpha$, where the gap of 0.348 contributes 0.043 to the eight-group mean.

\paragraph{Timing protocol.}
All timings exclude retrieval. We time MolReJev and Qwen3-4B-Text on one NVIDIA H200 in bfloat16, with one run per model and workload on the first 4,000 molecular, 1,000 DDI, and 1,000 USPTO records. MolReJev tokenizes prompts within the timed loop. Qwen3-4B-Text uses Hugging Face \texttt{generate} with FlashAttention-2 \citep{dao2024flashattention}, with timing starting after prompt tokenization and length sorting. MolecularGPT runs on Llama-2-7B-chat with its own prompt format and scoring protocol, and its timing includes prompt tokenization.

\begin{table}[!htbp]
\centering
\caption{Mean sequence lengths per record. For Qwen3-4B-Text, the input includes
the JSON answer prefix, and the output includes the answer text and its closing
quote.}
\label{tab:tokens}
\small
\begin{tabular}{@{}l ccc@{}}
\toprule
Task & Text baseline output & Text baseline input & MolReJev input \\
\midrule
Molecular properties      & 2.0  & 364.7  & 345.7  \\
DrugBank DDI (86 classes) & 18.6 & 2255.5 & 2319.5 \\
USPTO (50 classes)        & 7.1  & 981.0  & 975.3  \\
\bottomrule
\end{tabular}
\end{table}

\FloatBarrier
\subsection{Full-pool random forests}

Table~\ref{tab:rf-full} reports the labeled pool size for each group and compares random forests fitted to the full pool with MolReJev using four retrieved examples.

\begin{table}[!htbp]
\centering\small
\caption{Full-pool random forests and labeled pool sizes. RF fits all available pool records for each task. Pool sizes are means across tasks for binary groups and total record counts for the two multi-class groups. Binary entries report ROC-AUC, and multi-class entries report accuracy / macro-F1 on the full test sets. MolReJev uses four retrieved demonstrations. Bold marks the highest displayed score for each metric. Means weight the 16 binary groups equally.}
\label{tab:rf-full}
\setlength{\tabcolsep}{6pt}
\begin{tabular}{@{}lrrr@{}}
\toprule
Dataset (tasks) & Pool size & RF full & MolReJev \\
\midrule
\multicolumn{4}{@{}l}{\textit{Datasets used in training}} \\
FS-Mol (4507) & 399 & 0.715 & \textbf{0.782} \\
BACE (1) & 1{,}361 & \textbf{0.885} & 0.861 \\
BBBP (1) & 1{,}768 & \textbf{0.704} & 0.694 \\
HIV (1) & 37{,}008 & \textbf{0.831} & 0.821 \\
SIDER (27) & 1{,}284 & \textbf{0.644} & 0.640 \\
Tox21 (12) & 5{,}893 & 0.746 & \textbf{0.751} \\
BioSNAP (35) & 23 & \textbf{0.800} & 0.797 \\
AVIDa-hIL6 (16) & 13{,}727 & \textbf{0.951} & 0.926 \\
\multicolumn{4}{@{}l}{\textit{Multi-class datasets (accuracy / macro-F1)}} \\
DrugBank DDI (86 classes) & 134,265 & \textbf{0.876} / 0.763 & 0.843 / \textbf{0.779} \\
USPTO reactions (50 classes) & 9,000 & \textbf{0.949} / \textbf{0.948} & 0.912 / 0.911 \\
\multicolumn{4}{@{}l}{\textit{Datasets not used in training}} \\
FS-Mol held-out (157) & 268 & 0.717 & \textbf{0.760} \\
CYP450 (5) & 9{,}618 & \textbf{0.851} & 0.834 \\
ToxCast (530) & 2{,}274 & \textbf{0.628} & \textbf{0.628} \\
ClinTox (2) & 1{,}314 & 0.717 & \textbf{0.810} \\
BindingDB (152) & 129 & \textbf{0.923} & 0.916 \\
AVIDa-hIL6 held-out (15) & 14{,}052 & \textbf{0.954} & 0.944 \\
AVIDa-SARS-CoV-2 (12) & 2{,}990 & \textbf{0.750} & 0.726 \\
AVIDa-hTNF$\alpha$ (1) & 4{,}186 & \textbf{0.917} & 0.893 \\
\midrule
Mean, 16 ROC-AUC groups & --- & 0.796 & \textbf{0.799} \\
\bottomrule
\end{tabular}
\end{table}

\FloatBarrier
\section{Generalization and Training Overlap}\label{app:overlap}

A query and its reference pool are disjoint within each task, but the same structure may occur in several tasks. FS-Mol compounds can be screened in several assays, and the same antibody can be tested against different antigens. We measure how often a query occurs in the training records of another task and compare prediction on these queries with prediction on compounds absent from molecular training.

\subsection{Overlap with training records}
An evaluation query is \emph{shared} if its canonical SMILES (or raw sequence for antibodies) appears in any training record of a different task.
Task similarity is the cosine of TF-IDF vectors: character 3-gram vectors of the protein or antigen sequence for BioSNAP, BindingDB and AVIDa tasks (which share description text), and word-level vectors of the task description for all others.
For each shared query, we identify the most similar training task containing that molecule. Label agreement is the fraction of queries for which the two tasks assign the same label.

\paragraph{Prevalence.}
Table~\ref{tab:overlap-share} gives the counts.
Sharing is high for FS-Mol (75\,\%), FS-Mol held-out (77\,\%) and AVIDa-hIL6 (72--73\,\%), moderate for BBBP, SIDER, Tox21, BioSNAP, CYP450, ToxCast and ClinTox (36--57\,\%), and low for BACE (1\,\%), HIV (6\,\%) and BindingDB (13\,\%).
AVIDa-SARS-CoV-2 and AVIDa-hTNF$\alpha$ have no shared queries.
Where the most similar training task has text similarity $\ge 0.5$, labels agree 73\,\% (FS-Mol), 65\,\% (FS-Mol held-out) and 99.6--99.8\,\% (AVIDa-hIL6).

\begin{table}[!htbp]
\centering\small
\caption{Evaluation queries whose molecule (or antibody) appears in a training record of another task. ``Similar'' means the most similar such task has description (or sequence) similarity $\ge 0.5$; agreement is computed over those queries (BindingDB: 6 queries). Dataset sizes as in Table~\ref{tab:data}.}
\label{tab:overlap-share}
\begin{adjustbox}{max width=\linewidth}
\begin{tabular}{@{}lrrrrr@{}}
\toprule
Dataset & Queries & Shared & Shared (\%) & Similar & Label agreement (\%) \\
\midrule
\multicolumn{6}{l}{\emph{Used in training}} \\
FS-Mol & 211,145 & 159,130 & 75.4 & 112,464 & 73.1 \\
BACE & 152 & 2 & 1.3 & 0 & -- \\
BBBP & 197 & 111 & 56.3 & 0 & -- \\
HIV & 256 & 16 & 6.3 & 0 & -- \\
SIDER & 3,861 & 1,485 & 38.5 & 0 & -- \\
Tox21 & 3,072 & 1,289 & 42.0 & 0 & -- \\
BioSNAP & 309 & 175 & 56.6 & 0 & -- \\
AVIDa-hIL6 & 4,096 & 2,937 & 71.7 & 2,937 & 99.8 \\
\midrule
\multicolumn{6}{l}{\emph{Not used in training}} \\
FS-Mol held-out & 11,676 & 9,040 & 77.4 & 7,306 & 65.0 \\
CYP450 & 1,280 & 482 & 37.7 & 0 & -- \\
ToxCast & 83,342 & 41,299 & 49.6 & 0 & -- \\
ClinTox & 294 & 106 & 36.1 & 0 & -- \\
BindingDB & 4,996 & 662 & 13.3 & 6 & 50.0 \\
\shortstack[l]{AVIDa-hIL6\\held-out} & 3,840 & 2,812 & 73.2 & 2,812 & 99.6 \\
AVIDa-SARS-CoV-2 & 3,072 & 0 & 0.0 & 0 & -- \\
AVIDa-hTNF$\alpha$ & 256 & 0 & 0.0 & 0 & -- \\
\bottomrule
\end{tabular}
\end{adjustbox}
\end{table}

\subsection{Performance on shared and unshared queries} We compute ROC-AUC separately for shared and unshared queries within each task. Both subsets must contain at least three queries of each class. This requirement retains 535 of 4,507 FS-Mol training assays and 30 of 157 held-out assays, favouring assays with more queries; BACE, HIV, BioSNAP, AVIDa-SARS-CoV-2 and AVIDa-hTNF$\alpha$ retain no qualifying task. Table~\ref{tab:shared} reports three conditions: the untrained backbone with demonstrations (A), MolReJev (B), and MolReJev without demonstrations (C).

\begin{table}[!htbp]
\centering
\caption{ROC-AUC of the untrained backbone with demonstrations (A), MolReJev (B),
and MolReJev without demonstrations (C) on queries whose molecule is unshared (U)
or shared (S) with a training record of another task, averaged over the listed
tasks. Each task holds at least three queries of each class in both subsets.
The BBBP and ClinTox estimates each average over fewer than five tasks, and the shared subset of BindingDB holds 154 queries.}
\label{tab:shared}
\small
\setlength{\tabcolsep}{5pt}
\begin{adjustbox}{max width=\linewidth}
\begin{tabular}{@{}l c c cc cc cc@{}}
\toprule
 & & & \multicolumn{2}{c}{A: untrained} & \multicolumn{2}{c}{B: MolReJev} & \multicolumn{2}{c}{C: no demos} \\
\cmidrule(lr){4-5} \cmidrule(lr){6-7} \cmidrule(lr){8-9}
Dataset & Tasks & Queries U / S & U & S & U & S & U & S \\
\midrule
\multicolumn{9}{@{}l}{\textit{Used in training}} \\
FS-Mol               & 535 & 27,583 / 39,034 & 0.571 & 0.586 & 0.677 & 0.696 & 0.512 & 0.527 \\
BBBP                 & 1   & 86 / 111        & 0.725 & 0.596 & 0.778 & 0.633 & 0.533 & 0.501 \\
SIDER                & 26  & 2,288 / 1,430   & 0.600 & 0.577 & 0.645 & 0.665 & 0.523 & 0.658 \\
Tox21                & 10  & 1,484 / 1,076   & 0.605 & 0.665 & 0.710 & 0.769 & 0.619 & 0.650 \\
AVIDa-hIL6           & 7   & 449 / 1,343     & 0.935 & 0.929 & 0.976 & 0.945 & 0.511 & 0.554 \\
\addlinespace
\multicolumn{9}{@{}l}{\textit{Not used in training}} \\
FS-Mol held-out      & 30  & 1,531 / 1,983   & 0.627 & 0.602 & 0.739 & 0.727 & 0.544 & 0.540 \\
CYP450               & 5   & 798 / 482       & 0.805 & 0.810 & 0.839 & 0.858 & 0.566 & 0.660 \\
ToxCast              & 365 & 32,965 / 32,393 & 0.547 & 0.565 & 0.632 & 0.638 & 0.559 & 0.596 \\
ClinTox              & 2   & 188 / 106       & 0.822 & 0.574 & 0.832 & 0.749 & 0.490 & 0.574 \\
BindingDB            & 10  & 821 / 154       & 0.888 & 0.819 & 0.948 & 0.912 & 0.570 & 0.494 \\
AVIDa-hIL6 held-out  & 9   & 517 / 1,787     & 0.939 & 0.903 & 0.973 & 0.945 & 0.540 & 0.624 \\
\bottomrule
\end{tabular}
\end{adjustbox}
\end{table}

On both subsets, MolReJev exceeds the untrained backbone and benefits from demonstrations in all eleven groups. With demonstrations, the shared subset leads in five groups and the unshared subset in six. Without them, the shared subset leads in eight. The subsets differ in compounds and labels, so their scores reflect more than prior exposure.

\FloatBarrier
\section{Retrieval Variability and Class Balance}\label{app:retrieval}

Table~\ref{tab:retrieval} reports means and sample standard deviations across four retrieval draws at fixed weights and query sets. The standard deviations use the denominator $n-1$. About 4\,\% of AVIDa-hIL6 antibodies are positive, so four random examples seldom include a binder. Drawing two binders and two nonbinders improves ROC-AUC on both splits. Retrieving the four closest antibodies gives the highest ROC-AUC on both.

\section{Case Studies}\label{app:cases}

\begin{figure}[!ht]
\centering
\includegraphics[width=.95\linewidth]{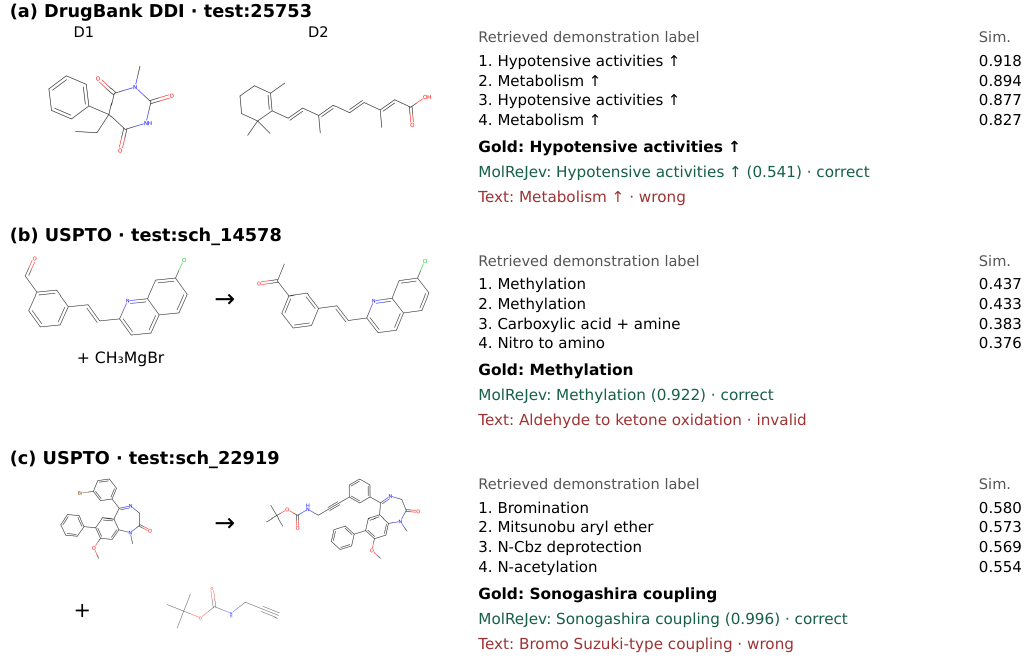}
\caption{Multiclass cases showing the queries, retrieved demonstrations and similarities, gold labels, and both predictions. Parentheses give MolReJev probabilities. Case 1 gives the full DDI labels. Reaction drawings omit solvents, spectators, and minor byproducts.}
\label{fig:cases}
\end{figure}

Figure~\ref{fig:cases} compares predictions from MolReJev and Qwen3-4B-Text on three multiclass queries. Both models receive the same four retrieved demonstrations for each query.

\paragraph{Case 1: Mixed demonstration labels.}
Two demonstrations carry the gold label ``D1 may increase the hypotensive activities of D2.'' The other two carry ``The metabolism of D2 can be increased when combined with D1.'' MolReJev gives these labels probabilities of 0.541 and 0.419, respectively. The text baseline predicts the metabolism type.

\paragraph{Case 2: A label outside the supplied set.}
MolReJev predicts the gold class ``Methylation'' with probability 0.922. The text baseline generates ``Aldehyde to ketone oxidation,'' a string outside the 50 candidate labels.

\paragraph{Case 3: A class absent from the examples.}
MolReJev assigns 0.996 probability to the gold class, ``Sonogashira coupling,'' even though none of the demonstrations belong to that class. The text baseline predicts ``Bromo Suzuki-type coupling.'' The query contains the alkyne--halide coupling pattern associated with the gold class.

\end{document}